# hmBlogs: A big general Persian corpus

**Hamzeh Motahari Khansari, Mehrnoush Shamsfard**

NLP Research Laboratory, Computer Science and Engineering Dept., Shahid Beheshti University, Tehran, Iran

{h_motahari, m-shams}@sbu.ac.ir

**Abstract**

This paper introduces the hmBlogs corpus for Persian, as a low resource language. This corpus has been prepared based on a collection of nearly 20 million blog posts over a period of about 15 years from a space of Persian blogs and includes more than 6.8 billion tokens. It can be claimed that this corpus is currently the largest Persian corpus that has been prepared independently for the Persian language. This corpus is presented in both raw and preprocessed forms, and based on the preprocessed corpus some word embedding models are produced. By the provided models, the hmBlogs is compared with some of the most important corpora available in Persian, and the results show the superiority of the hmBlogs corpus over the others. These evaluations also present the importance and effects of corpora, evaluation datasets, model production methods, different hyperparameters and even the evaluation methods. In addition to evaluating the corpus and its produced language models, this research also presents a semantic analogy dataset.

## 1- Introduction

Language corpora are basic resources in natural language processing (NLP). These corpora range from small to very big (including billions of words). They can include just raw text or have some meta-data such as tags and annotations. As a general rule, it can be said that a larger corpus leads to a more useful corpus. This is especially the case when the main reliance is on statistical methods.

Persian (Farsi) has an especially wide range of speakers across the world, specifically in Iran, Tajikistan and Afghanistan and is used as either a native or second language. It is, therefore, especially important to develop the necessary NLP tools and resources for this language. A major downfall of Persian as a low resource language is its lack of sufficiently large, covered and up-to-date corpora.

In recent years, the use of language models based on word embeddings has become a common phenomenon. These models are not only crucial and practical on their own, but also have deep neural networks applications. These models are produced by the corpora and their quality strongly depends to the used corpora. A basic requirement for a corpus is its ability to represent the language adequately and incorporate the necessary genres and domains. The Persian language, however, lacks a widely available corpus with such features.

The hmBlogs corpus is presented in this research as a large, accessible, public and diverse corpus of various language genres, and some of the word embedding models prepared by the hmBlogs corpus are also included. A set of analogy dataset prepared to evaluate this corpus is provided alongside, and the hmBlogs corpus is compared and evaluated based on the present dataset and two other datasets with some of the most important available Persian corpora.

## 2- Related work

Among the various corpora (with different dimensions and areas) prepared for the Persian language, some are private and publicly unavailable and others are either publicly available or accessible to researchers. It can be claimed that the volume and variety of corpora available for the Persian language is much less than that available for other languages such as English, which is a laboratory language.

Corpora can be obtained from different news websites, newspapers and magazines, books, blogs and tweets, and etc. Due to a lack of Persian content on the Internet in the past years, older sources, such as the Hamshahri corpus (AleAhmad et al., 2009), rely less on Internet-created data. Newer corpora, such as the MirasText (Sabeti et al., 2018), on the other hand, are made only from the Persian content found on the Internet; and, even the presence of snippets is due to the publication of those snippets on the Internet.

The Hamshahri corpus (AleAhmad et al., 2009) is one of the oldest Persian corpora and was founded on the compilation of news and articles of the Hamshahri newspaper, one of the most famous newspapers in Iran. It contains more than 166 thousand news and more than 63 million words. Its content is dated between the years 1996 and 2002. Articles in this corpus are labeled in 12 main thematic categories. One problem with this corpus is its lack of up-to-date content, which means that it fails to include certain important language and content changes such as changes in language and linguistic framing, semantic drifts and appearing new names and events. Due to its news nature, it is also devoid of the different types of language uses common among Persian speakers. An advantage of this corpus, however, is its accuracy and lack of typographical errors of the compiled text.

Peykareh (Bijankhan et al., 2011) is another early popular Persian corpus for NLP applications. This corpus contains 100 million words from a collection of news, web sites, and written sources. Peykareh texts range from the year 1978 to 2003. An important feature of this corpus from an NLP stance is its POS tagging, which is done semi-automatically. Of course, only about one tenth of this corpus has been labeled. Based on the word embedding training and the fact that its latest texts were compiled before 2004, Peykareh is a corpus of old texts. Furthermore, its size may not be large enough to make appropriate word embedding models.

Another important resource used in NLP field is Wikipedia. The Persian Wikipedia (FaWiki) was launched in December 2003 and contained more than 775,000 articles by March 2021 ("Persian Wikipedia," 2021). Persian is currently the 19th language on Wikipedia (*List of Wikipedias*, n.d.). Like the Hamshahri corpus, the FaWiki has an official language that cannot cover all language types. Furthermore, due to its encyclopedic nature, a large number of Wikipedia articles are too specialized for Persian speakers and everyday usage. For example, the vast majority of Persian speakers do not know the capital, or currency of many countries. Or many people are unfamiliar with the names of drugs, programming languages or specialized topics in astronomy or anatomy. The existence of such articles and the preparation of word embedding based on such can cause incorrect bias and a disconnect between the obtained model and the real language of the people. Of course, an important advantage of Wikipedia compared to others is its constant updates and continuous expansion.

IrBlogs (AleAhmad et al., 2016) is a corpus that has been created by crawling Persian blogs. Blogs are one of the most important resources for making corpora and are a valuable and rich source of a variety of texts due to a diversity of authors and topics. IrBlogs is the result of the crawl of more than 564 thousand blogs and includes nearly 5 million posts. It contains posts collected from 2002 to 2013 to study the Persian blogging space. This corpus also includes posts that are not in Persian. The existence of posts with other languages impairs the preparation of a suitable language model. Also, the newest texts of irBlogs are about 7 years old and do not include the most recent changes such as new names and events.

Another text corpus prepared for the Persian language is the MirasText (Sabeti et al., 2018). When published in 2018, the MirasText was considered to be the largest corpus available in Persian. It contains about 2.8 million documents and about 1.4 billion words collected from around 250 websites, of which at least 150 sites of them are news websites. In addition to plain text, this corpus provides information, such as titles, descriptions, and keywords for its documents.

The Persian Raw Text (PRT) (*Persiannlp/Persian-Raw-Text,* 2020/2021) is a corpus provided by putting together a variety of Persian corpora (including MirasText and FaWiki) and is not produced independently. Also, no special processing has been performed on it to detect any duplication. The main part of PRT is obtained from the Persian section of the Common Crawl project (*Common Crawl*, n.d.). Common Crawl is a web crawl project that crawls and collects resources available on the web in any language, including some in Persian. PRT is larger than hmBlogs in size, but as shown in the evaluation section, it scores lower than hmBlogs.

## 3- Corpus Crawl

The crawl process of hmBlogs is described in this section.

### 3-1- Available resources for preparing a corpus

A large and comprehensive corpus requires a high volume and variety of texts to cover the maximum space of the Persian language used by its speakers.

The Internet and its content is one of the first and foremost resources available for free. Main text sources for Persian content on the Internet can be divided into four general groups:

- News sources: News websites have a long history of content production in the Persian language. These sources have a formal language and focus only on news events. Texts in these sources are usually produced with a similar procedure, structure and style and have a specific vocabulary. For example, it is uncommon to find works of fiction, poetry, folk or colloquial texts in the news. While the heavy replication and copying of news in the media negatively affects the quality of the corpus, removing duplicated data will not necessarily be easy. Although duplication is a problem that can be found in other Internet sources, duplication in the news media is an even more serious problem. The influx of daily news can lead to redundancy.
- Blogs: Blogs make up for one of the most important textual sources of Persian content on the Internet. Blog popularity among Persian speakers had placed Persian among the top ten popular languages among bloggers in the world (Megerdoomian & Hadjarian, 2010). Bloggers cover a wide range of different people with different social and educational levels, and blogs contain a wide variety of content from different genres, styles and areas. The blog space is especially important as each has its own specific characteristics. The rise of social networks, however, jeopardized the presence of blogs.
- Social networks: Social networks became the next repository for a large amount of Persian content. For example, the Telegram Messenger (one of the most popular instant messengers in Iran) has millions of Persian subscribers and is considered to be one of the most important textual sources of Persian content. The main problem with data crawling on social networks is the complexity of finding adequate texts. Many groups, channels, and social pages are not publicly available or known, and some of them are not even indexed by an Internet search engine. In addition, due to their shorter lived life than the previous two sources, social networks cover a limited amount of Persian texts. Another problem is that some of the specific styles of these networks affect the production of text in them. For example, the use of hashtags is very common in social networks, especially in posts that contain advertisement. Word count limitation, such as those observed with Twitter, is another example of how the prevalence of a certain style of writing does not necessarily conform to the normal method of writing texts.

- Other websites: In addition to the resources mentioned above, there is a variety of websites that cover a wide range of texts. Crawling these websites can also be valuable for corpus production. However, the structural diversity of these websites makes it difficult to extract pure text from the unrelated elements of a web page. Another problem is the lack of a specific criterion to ensure the obtained corpus is a balanced corpus. Due to the unavailability of a balanced list of such sites, the vast data volume, and the inability to estimate the amount of data needed to be removed from each website to ensure a relatively balanced corpus, the use of these websites may not be very practical or useful.

### 3-2- Choosing Blogfa as a source of corpus

Considering the issues raised in the previous section, it can be said that due to the variety of language types, different writing styles, its long history, wider range of posts, and simplicity in crawl, the use of blogs is a good option for providing corpus. Challenges facing crawling blogs include:

- Finding the address of blogs and preparing a list of crawling
- Finding the direct address of each blog post
- Locating the post from among various other elements on each post page (header, footer, ads, duplicated sections related to the blog template, and etc.)
- Extracting the pure text of the post without disrupting the text structure and removing other non-text elements (such as photos, hyperlinks, and etc.) from the post.

These challenges make it difficult to crawl diverse blogs with different structures. Therefore, the tendency is to crawl blogs that use relatively uniform structures so the crawler can be developed with the least cost and most accuracy.

Fortunately, most bloggers use a limited number of blog hosting services such as Blogger, WordPress and Medium, which are some of the most popular blog hosting services in the world. Many Persian-speaking bloggers also use free Persian blog hosting services. The Blogfa hosting service is especially popular with many Persian-speaking users. For example, about 70% of crawled blogs in the irBlogs corpus are hosted by Blogfa. So it can be said that crawling Blogfa alone covers the dominant part of the Persian blog space. In addition, another important advantage of Blogfa is its direct post addressing protocol, which provides access to all posts on a blog using a simple algorithm. For this reason, Blogfa was chosen as the source of the corpus for the crawl, and the Python language was used to develop the crawler.

The first challenge in crawling Blogfa was finding a comprehensive list of blogs (i.e. blogs' URLs). But because of the absence of public and free lists, the list was obtained in the following manner:

- Some lists of recently updated or categorized blogs can be found on the Blogfa homepage, which includes only a small portion of Blogfa blogs. Crawling these pages helped us find about 20,000 blog addresses.
- About 20,000 addresses were extracted from a sample dataset obtained from the Yooz Iranian search engine project.
- About 200 addresses were added manually.

- The most important source for finding blog addresses is through extracting the addresses of new blogs from crawled blogs. A common practice among bloggers is putting the links of other blogs on their own blogs. This practice is often a two-way practice, in which a so-called link exchange takes place. In this way, new blogs are introduced and reach a larger audience. About 670,000 new blog addresses discovered from this source.
- In addition, to ensure maximum coverage of all language types and blogs, a list of Persian words was prepared after the first phase of crawl (see 3.3 section) to find the blogs most probably missed. In this regard, a list of nouns of FarsNet (Shamsfard, Hesabi, et al., 2010) that were not multi-token (without spaces (blanks) and half-spaces[1]) was selected, and a maximum of 10 blogs related to each word were searched in the Bing search engine. The result was the discovery of about 30,000 new blogs. Also the crawl of these 30,000 new URLs led to the discovery of other new blogs that were identifiable based on the link exchange behavior.

### 3-3- Crawl of Posts

Addresses of posts in Blogfa follow a simple contract. Each post can be accessed directly by the following address template:

http://XXXXX.blogfa.com/post/POST_NUMBER

Where XXXXX refers to a unique blog name and POST_NUMBER is an integer identifier.

Cases where a specific post was deleted by the author over time receive a 404 error code and simply are ignored.

During the hmBlogs corpus creation, blogs and their posts were crawled somewhat randomly to avoid bias as much as possible. The 40,000 addresses extracted initially as the primary seed were also sorted randomly. Then a blog was selected from the head of the list of discovered but not crawled addresses, and its home page, containing the latest blog posts, was crawled. Accordingly, the last (maximum) post number was extracted based on the post address pattern in Blogfa, and an estimate of the total number of posts published in this blog was discovered. A total of 100 posts were selected randomly from among the posts, which might have even included previous deleted posts by the author, for crawling, and their addresses were created and crawled directly. To prevent the addition of duplicated material on the corpus, the first and the last posts of each blog were excluded from the process. The disregard is mainly due to their common themes, such as greeting the audience or bidding them farewell. Once completed, the crawler moves on to the next blog that is on the crawl queue. Any new discovered Blogfa blog addresses (from the blog's home page) will, in the meantime, append to the crawl list.

Furthermore, the presence of deleted posts or blogs with less than a total of 100 posts brings the number of crawled posts from each blog to practically less than 100 in many cases. As observed in Table 1, deleting empty, encrypted and duplicated posts left an average 47 posts per blog.

Using the multi-threaded method can help speed up the crawl process. In this method, as a result, the consecutive obtained posts are not necessarily from one blog. This has, however, improved the random order of blog posts in hmBlogs.

---

[1] Half-Space character, more officially name Zero-Width Non-Joiner (ZWNJ) character is a type of space in Persian writing. This type of spacing indicates space between two or more parts of a word, while not causing the word to be considered as a multi-token word and keeps it as a single-token word. This type of spacing is not adhered strictly by Persian speakers and sometimes replaced by using regular space character or even in sometimes it is omitted.

HmBlogs was crawled in two phases. The first phase took about two months to complete the crawl process, and the second phase (about one year later) lasted only two weeks. To ensure better coverage of possible topics, genres and styles, the crawl process was updated and completed in the second phase by about 30,000 new discovered blog addresses. Another advantage of the second phase was the addition of new posts related to the Covid-19 pandemic.

Details are listed in the Table 1 table:

Table 1. HmBlogs crawl details.

| **Description** | **Value** |
|---|---|
| Number of discovered Blogfa blog addresses | 745,153 |
| Number of Blogfa blogs successfully crawled | 438,624[2] |
| Number of crawled posts (Not empty, not encrypted[3] and not exactly duplicated posts) | 20,538,571 |
| Average number of posts per blog | 46.83 |
| Number of tokens (delimited by space character) | 6,962,351,774 |
| Average number of tokens per post | 338.99 |

**3-4- Data Cleaning and Outputs**

The noise and errors present in the primary raw crawled corpus reduce the quality of the corpus in many NLP tasks and can be eliminated using preprocessing. The output of the corpus based on these preprocessings is presented in both raw and cleaned modes[4]. The word embedding models for the corpus evaluation was prepared by the cleaned mode of the corpus. The preprocessings conducted on hmBlogs are as follow:

- Deleting posts in other languages: A significant number of posts that were written in the Persian alphabet were not actually in Persian. Because the Persian alphabet has a lot in common with the alphabet of languages such as Arabic, Turkish or Kurdish, a Persian language detector was developed to determine the dominant language of a post. A list of common words in Persian, based on the FarsNet single-token words and other selected correct Persian texts, was initially prepared as the reference list. Then, a unigram list was extracted for each post and the percentage of the intersection of the unique words in that post and the reference list was calculated. If the common words of the post exceeded the predefined threshold, the post was assumed a Persian post. This study was performed with two intersection percentages, 50% and 75%, and the word embedding models were made based on their outputs. The results showed that the model made based on the corpus with a 50% separation threshold was better than the model with a 75% threshold. After selecting the 50% criteria, the non-Persian posts were separated and deleted accordingly. Thus, posts that were in languages other than Persian were removed from both outputs of the corpus (raw and cleaned). In addition to non-Persian posts, posts with severe typographical errors or very rare words were also removed in this process, which helped reduce the overall corpus noise.

[2] The long distance between the number of discovered addresses and the number of successfully crawled blogs is due to the deletion of many of the blogs during years.
[3] Blogfa allows bloggers to encrypt some of their posts, so only certain people they wants can see the content of the post. Naturally, the content of these posts was not readable by the crawler and was removed from the output.
[4] Both raw and cleaned outputs of hmBlogs can be accessed at http://fs.nlp.sbu.ac.ir/members/motahari/metr/papers/hmBlogs/

- Deleting empty or almost empty posts: Some posts with no text or very short and relatively duplicated texts were removed from both corpus outputs. These posts usually only contain images or hyperlinks to other pages or resources. Also the creation of empty posts may be due to the crawler's inability to properly extract a post text from the HTML code of the page. This has been observed for users who have changed the style of their pages from the default Blogfa items.
- Converting non-standard characters to standard Persian characters in the cleaned output: One of the problems with electronic documents in Persian is the existence of some non-standard characters. These characters are sometimes mistakenly replaced by characters from other languages (mainly Arabic). A common example is the Arabic ‘ي’ and ‘ك’, which are sometimes used instead of the Persian ‘ی’ and ‘ک’. These characters have an almost exact appearance as their Persian counterparts, but their different codes causes the computer system to assume they are different. Another aspect affecting Persian blogs is the use of characters that make the text look fancy, such as the non-Persian character ‘ב’ instead of the letter ‘د’ (equivalent to D in English). In fact, some Persian bloggers have even used such characters to make their texts look nicer.
- In the cleaned output of the corpus, all characters outside the range of numbers, space, half-space, new line character ('\n'), Persian alphabet characters, English alphabet characters and a set of main punctuation characters in English and Persian (including the characters '.', '!', '?', ' ,'؟ ؛' ,'،' ,':' ,';' ,',') have been deleted.
- Uniformity of numbers: There are millions of different numbers in crawled posts that can create noise in word embeddings rather than being useful. For this reason, all numbers are replaced by a default number[5] in the cleaned output of the corpus.
- Correcting words with repetitive characters: Similar to English writers, Persian writers may try to show emphasis by repeating characters in a word. For example, the word ‘خوب’ (‘Good’ in English) may be written as ‘خووووووب’ (like ‘Gooooooood’). Since it is very rare that a Persian word has more than two repeated letters right after each other, all the characters that have been repeated more than twice in a row have been replaced with the one equivalent character in the cleaned output of the corpus.
- Deleting all non-Persian sentences: The sentences that appear in the posts and do not contain any Persian alphabet characters were removed from the cleaned output of the corpus, as these sentences are not useful for word embedding training.
- Converting all half-spaces to space: Due to the absence of a uniform procedure between Persian speakers in using the half-space character, and in many cases it is common that the space character is used instead of them or even deleting them at all, in order to reduce noise in the construction of word embedding models, all half-spaces are replaced by space in the cleaned corpus output to reduce noise in the construction of word embedding models.

[5] The number 1399 is used, which indicates the current year in the Iranian calendar (equivalent to 2020-2021 in Georgian calendar).

Output details in both raw and cleaned corpora after preprocessing are given in the Table 2:

Table 2. HmBlogs details (as raw and cleaned corpus).

| **Description** | **Value** |
|---|---|
| Number of posts | 19,942,663 |
| Number of tokens in raw corpus (delimited by space character) | 6,832,123,341 |
| Number of tokens in cleaned corpus (delimited by space character and any punctuation excluded) | 6,782,035,655 |
| Number of types[6] in cleaned corpus (punctuations excluded) | 12,642,391 |
| Average number of tokens per post (in cleaned corpus) | 340.08 |

Another problem evident in the corpus is the presence of typos and the carelessness of authors for following the correct Persian orthography. For example, in many cases some letters and words are mistakenly attached to each other as in the phrase 'خواهران و برادرانش' ('his sisters and brothers'), in which the letter 'و' ('and') is attached mistakenly to 'برادرانش' ('his brothers'). As a result, the three-token phrase is changed to a two-token phrase 'خواهران وبرادرانش'. A tokenization tool such as Step-1 tokenizer (Shamsfard, Jafari, et al., 2010) or a spelling and grammar checker tool, like Negar (*Negar*, n.d.), may be helpful in correcting such errors of the corpus. Problems that can arise in such situations include:

- Running these tools on a large corpus such as hmBlogs is too time consuming.
- Generally these tools are not very accurate or reliable in Persian and can lead to mistakes.
- These tools, at least in Persian, can malfunction when faced with informal or colloquial text because they are usually programmed for formal texts.

Due to the mentioned challenges, these types of preprocessings and corrections were ignored for hmBlogs.

---

[6] Unique tokens

## 4 - Corpus Evaluation

### 4-1 - Preparation of Corpora for Evaluation

As mentioned, the main purpose of producing hmBlogs is helping to prepare perfect language models. The language models produced on this basis were used in the corpus evaluation. To ensure the quality of the hmBlogs, we compared the corpora of IrBlogs, PRT and FaWiki with that of hmBlogs.

The closest and most relevant comparable corpus to hmBlogs is irBlogs, as it too is formed on the data obtained from Persian blogs. This corpus is provided in the form of many XML files. In order to compare and prepare word embeddings, it was necessary to extract the text of the posts from XML files. Also, for fair comparison, the entire preprocessing for hmBlogs was applied to irBlogs at the internal content level of its posts. During the pure text extraction process, a small part of the posts were found to have errors in their XML formats and were not parsed properly as a result. A total of 4,977,920 posts were extracted from the 4,983,359 original irBlogs posts (about 99.9%). The Table 3 shows the details of the preprocessing on irBlogs:

Table 3. Details of preprocessed irBlogs.

| **Description** | **Value** |
|---|---|
| Number of extracted posts from XML files | 4,977,920 |
| Number of empty posts | 629,681 |
| Number of nearly empty posts | 478 |
| Number of valid and successfully preprocessed posts | 4,347,761 |
| Number of tokens in the preprocessed corpus | 881,155,421 |
| Number of types in the preprocessed corpus | 3,701,875 |
| Average number of tokens per post | 202.67 |

Another corpus that are usually used for comparison and evaluation are the Wikipedia based corpora. In this regard, the pure text of the Persian wiki (FaWiki) was extracted from (*Persiannlp/Persian-Raw-Text*, 2020/2021)[7]. Since this corpus is in plain text and does not have the post feature, a feature observed in blogs, preprocessings in hmBlogs were made on it without considering this concept. The result was a corpus with 87,048,240 tokens and 831,019 types (delimited by space character and any punctuation is excluded).

Since irBlogs and FaWiki, as compared with hmBlogs are both smaller in size, the Persian Raw Text (PRT) corpus was selected as a bigger corpus than hmBlogs to evaluate whether or not just increasing the corpus size can lead to better models. Since this corpus is in plain text and without posts, the hmBlogs preprocessings were made on it without considering the post concept. The result was a corpus with 7,970,115,160 tokens and 8,257,949 types (delimited by space character and any punctuation is excluded). The interesting thing is that although the number of tokens in this corpus is one billion more than that of hmBlogs, its vocabulary size (types) is about 65% of that of hmBlogs.

### 4-2- Preparation of Datasets for Experiment

As mentioned, hmBlogs evaluation is based on the evaluation of word embedding models produced from hmBlogs in comparison with models produced from other corpora. For purposes of this research, we selected the analogy test and semantic relatedness methods to evaluate the word embedding models.

In short, an analogy test is based on a successful guess of the fourth related word based on its predecessor words. For example for ‘Family’ relation, ‘Man’ is to ‘Woman’ as ‘Boy’ is to ‘Girl’. If a model can successfully guess the word ‘Girl’ as the fourth word, this shows that the model is working properly. In semantic relatedness, a similarity value is defined between two words to show their relatedness. For example it can be said that the relativity between ‘Airplane’ and ‘Passenger’ is 5, between ‘Airplane’ and ‘Road’ is 2 and between ‘Airplane’ and ‘Sun’ is 0. If a model can guess the similarity value as much as same as humans do, this shows that the model is a better model compared to others.

Each of these evaluation methods requires their own datasets. The following is an explanation of the three datasets used for this purpose.

#### 4-2-1- Google Analogy Test Set (GATS):

This dataset, which is introduced by (Mikolov et al., 2013) and is well known among researchers, was selected as the first option for evaluation. Naturally, since the data are in English, it had to be translated into Persian first. The translation was initially done automatically by the Google translator and then edited manually. There are several problems with this path (as some of them were reported by (Hadifar & Momtazi, 2018)):

- One problem with translating the dataset is that all its English words are single token words. While in Persian translation, many of them become two-token (or even more) words. For example, the word ‘simpler’ becomes a two-token word (‘ساده تر’), or ‘vanished’ is translated to ‘ناپدید شد’, which is also a two-token word. In particular, in translating into Persian, many GATS verbs have more than single-token words. Words that is translated into more than single-token words generates an Out of Vocabulary (OOV) error in the evaluation and made the evaluation of the corpora ineffective. This is because word embedding models are trained based on splitting words on solely space characters and they don’t consider multi-token words. This mainly arises from challenges in tokenization in Persian and beyond the scope of this research and should be addressed separately.
- A second problem with translating the dataset is that Persian writers do not follow the same exact orthography for many words. For example, among Persian speakers, the word ‘debug’ has been understood and accepted in three forms: ‘اشکال‌زدایی کردن’, ‘اشکالزدایی کردن’ and ‘اشکال زدایی کردن’. Such cases make it difficult to map an English word to a specific word in Persian.

[7] Dumped on March 2020

- A third problem to this phenomenon is the lack of equivalent Persian words for some English words. For example, English distinguishes the third person based on gender, while the third person pronoun in Persian does not have a specific gender, so the ‘She / He’ is converted into only ‘او’. There are also cases where the Persian equivalent of a word in English is several words. For example, uncle in Persian can be translated as ‘عمو’ (brother of father) and ‘دایی’ (brother of mother).
- Another problem is that some categories in GATS do not have exact equivalents in Persian. For example, the verb ‘come’ in English can be written in two forms, ‘come’ and ‘comes’, whereas in Persian, verbs have 6 forms: ‘می‌آیید’, ‘می‌آییم’, ‘می‌آید’, ‘می‌آیی’, ‘می‌آیم’, and ‘می‌آیند’. So it is not possible to translate the Plural-Verbs category of GATS directly into Persian. Accurate translation becomes near impossible in such cases.
- Some data in GATS are not in accordance with the culture and context of Persian texts. For example, due to the Persian nature of the corpus as opposed to English, a lot more emphasis is placed on evaluating the provincial centers in Iran than on learning names of the United States capital cities.
- Translating and mapping polysemous words and choosing the appropriate Persian word in senses that contain more than one word is also very challenging. This challenge is especially evident in the automatic translation from English to Persian, which needs to determine the correct meaning before translating begins. Because the words in analogy questions are not used in sentences, word-sense disambiguation (WSD) techniques are not practical. Furthermore, the complexities and ambiguities of manual translation are challenging even for the human expert.

As observed in the above stated challenges, the literal translation of GATS did not deliver accurate results, both in terms of words or even categories. However, the model output evaluations based on this inaccurate translation showed the superiority of hmBlogs. But since this translated dataset was deemed inaccurate, all corpora evaluations based on the dataset were considered weak, and the translated dataset was, therefore, discarded.

As the next step, a GATS-like dataset was produced for Persian by (Zahedi et al., 2018) and provided to this research. Similar to GATS, the produced dataset fall under both semantic and syntactic categories, with a total of 15 categories (as opposed to the 14 categories of GATS). This dataset also has the number of 31,332 analogy questions (as opposed to the 19,544 questions in GATS), which makes it superior to GATS in these respects. Some categories in Persian, which are semantically similar to English (such as Country-Capital), are almost duplicated. Also some categories, such as Province-Capital, which is similar to the City-in-State category in GATS, are in accordance with the Persian and Iranian culture, and other categories that did not have a meaning in Persian are replaced with more accurate Persian counterparts. It should be noted that since the half-spaces were converted to space character in the preprocessed corpora, all half-space characters were removed from this dataset to help make a more accurate assessment.

Examples[8] of this dataset are presented in Table 4:

Table 4. Some samples of Persian equivalent of GATS (selected from (Zahedi et al., 2018)).

| Relation | Word1 | Word2 | Word3 | Word4 |
|---|---|---|---|---|
| Adjective-Adverb | سریع (Fast) | سریعا (Fast) | کامل (Complete) | کاملا (Completely) |
| Country-Capital | فرانسه (France) | پاریس (Paris) | عمان (Oman) | مسقط (Muscat) |

[8] Samples are selected as they can be comprehensible for audiences are unfamiliar with Persian.

| | | | | |
|---|---|---|---|---|
| 1st Person | آمدم (I came) | آمدیم (We came) | گفتم (I said) | گفتیم (We said) |
| Nationality | آلمان (Germany) | آلمانی (German) | مالزی (Malaysia) | مالزیایی (Malaysians) |
| Noun-Adverb | ماه (Month) | ماهانه (Monthly) | شب (Night) | شبانه (Nightly) |
| Infinitive-Past | پریدن (To Jump) | پرید (Jumped) | رفتن (To Go) | رفت (Went) |
| Singular-Plural | کتاب (Book) | کتاب‌ها (Books) | مدرسه (School) | مدارس (Schools) |
| Infinitive-Present | رسیدن (To Arrive) | می‌رسد (Arrive) | نوشتن (To Write) | می‌نویسد (Write) |

#### 4-2-2 FarsNet Analogy Test Set (FATS)

An analogy dataset, produced specifically for hmBlogs assessment is FATS (FarsNet Analogy Test Set). FATS was found on the basis of (Mahmoudi, 2020), with the difference that instead of using a manual extraction, an automatic semantic relation extraction from FarsNet is used. FarsNet is the largest and most accurate equivalent for WordNet (Fellbaum, 1998) in Persian and has over 100,000 lexical entries organized into more than 40,000 synsets, and nearly 100,000 semantic relations established between its synsets. Analogy questions in FATS are generated automatically based on the mentioned semantic relations in FarsNet, and word embedding models are evaluated on the basis of how well they can guess these types of relations. To prepare the dataset, all semantic relations in FarsNet[9] were extracted, and the relations that their participated words are single-token words[10] were separated, and as same as GATS questions structure, the dataset was prepared. The big problem was that the number of instances in some relationships was so many[11], as the volume of the final dataset was equal to 121,792,416 analogy questions. Evaluating this dataset with multiple models was both time consuming and beyond the available hardware capabilities of this research. Therefore, the dataset was programmed to generate a maximum of only 100, 250 and 500 items from each category of the relations, respectively. A total of 135,166, 609,904 and 1,957,314 questions were generated as a result, respectively. Evidently, these numbers of questions are not comparable with those of GATS (19,544 questions) or (Zahedi et al., 2018) (31,332 questions). In order to ensure maximum fairness in the generation of questions, the relations were randomly selected. Therefore, the questions generated in each mode (FATS-100, FATS-250 or FATS-500) were not, necessarily, common with the other modes. The number of categories generated in FATS is 33, and all of them are semantic relations[12].

Examples[13] of the questions in FATS can be observed in Table 5:

Table 5. Some samples of FATS.

| Relation | Word1 | Word2 | Word3 | Word4 |
|---|---|---|---|---|

[9] Except "Related-To" relation which has a very wide meaning and cannot be used for analogy question generation.
[10] And also do not have half-space.
[11] Examples: 10,568 Hypernym relations and 1023 Hyponym relations.
[12] The generated datasets are available in http://fs.nlp.sbu.ac.ir/members/motahari/metr/papers/hmBlogs/dataset/.
[13] Samples are selected, as they can be comprehensible for audiences unfamiliar with Persian.

| | | | | |
|---|---|---|---|---|
| Agent | نوازنده (Player) | نواختن (Play) | باد (Wind) | وزیدن (Blow) |
| Antonym | خروج (Exit) | ورود (Entrance) | شیرینی (Sweetness) | تلخی (Bitterness) |
| Has-Salient defining feature | فلفل (Pepper) | تند (Hot) | هیولا (Monster) | ترسناک (Scary) |
| Hypernym | پفک (Cheetos) | تنقلات (Junk Food) | حسابداری (Accounting) | علم (Science) |
| Hyponym | سبزی (Vegetable) | اسفناج (Spinach) | دام (Trap) | تور (Net) |
| Instance hypernym | ترکیه (Turkey) | کشور (Country) | بیروت (Beirut) | پایتخت (Capital) |
| Is-Agent-of | دونده (Runner) | دویدن (Run) | جونده (Rodent) | جویدن (Gnaw) |
| Part meronym | آسیا (Asia) | قطر (Qatar) | همبرگر (Hamburger) | گوشت (Meat) |

### 4-2-3- Semantic Relatedness Dataset

As mentioned before, another evaluation method used for word embedding models is the semantic relatedness assessment between words. For this purpose, the SemEval2017-Task2 dataset (Camacho-Collados et al., 2017) was used. This dataset contains 500 pairs of nouns, which have been rated according to their similarities by humans. Of course, the dataset also includes multi-token words, which cannot be used to evaluate the models of this experiment and are all based on single-token words. Therefore, all multi-token words were removed from this list, and, finally, 354 pairs of single-token words were tested. Also, since half-spaces were replaced by space character in the preprocessings of the corpora, all half-space characters were removed from the dataset to help make a more accurate assessment.

## 4-3- Evaluation and Results

To evaluate hmBlogs, word embedding models needed to be trained on it and on other comparable corpora in different ways. Word embeddings are prepared using the following four methods:

- Word2Vec (Skip-gram) (Mikolov et al., 2013)
- Word2Vec (CBOW) (Mikolov et al., 2013)
- GloVe (Pennington et al., 2014)
- FastText (Skip-gram) (Bojanowski et al., 2017)

There are various hyperparameters for training word embedding models that require multiple experiments to find the optimal ones. For this purpose and to study the effect of hyperparameters, Word2Vec was used initially to train several models with different hyperparameters based on hmBlogs and irBlogs, evaluate the generated models, and then select the best hyperparameters to train the models based on GloVe and FastText in the next step[14]. The experiments are further discussed in the following sections.

### 4-3-1- Experiment on the Persian equivalent of GATS

As mentioned above, in order to achieve optimal hyperparameters, models with different hyperparameters were trained initially, and their results were evaluated. The Table 6 shows the results of the evaluation (Top-5) and comparison of the produced models with different hyperparameters based on Word2Vec for both hmBlogs and irBlogs. Gensim (Řehůřek & Sojka, 2010) is used to generate the models and work with them, and (Bairathi, 2017/2018) is used to evaluate and find the accuracy percentages.

Table 6. Accuracy percentage results on Word2Vec training for both hmBlogs and irBlogs with different hyperparameters.

| Dimension | Win. Size | Min. Count | SG or CBOW | hmBlogs | irBlogs |
|---|---|---|---|---|---|
| 400 | 5 | 2000 | SG | 43.291 | 20.640 |
| 400 | 5 | 2000 | CBOW | 42.643 | 20.407 |
| 400 | 5 | 1000 | SG | 48.066 | 29.829 |
| 400 | 5 | 1000 | CBOW | 47.536 | 29.452 |
| 300 | 5 | 100 | SG | 52.694 | 44.753 |
| 400 | 5 | 100 | SG | 53.514 | 45.503 |
| 400 | 10 | 100 | SG | 53.935 | 45.822 |
| 400 | 5 | 5 | SG | 51.392 | 42.736 |

According to the items mentioned in the Table 6, the following points can be made:

- In all cases, hmBlogs is far ahead of irBlogs.
- In all cases, the performance of Skip-gram and CBOW was close to each other, and overall Skip-gram was superior.
- Dimension reduction from 400 to 300 had little negative effect on the results.
- Increasing window size from 5 to 10 had little positive effect on results.
- Decreasing the minimum count had an obvious positive effect on the results. But this reduction had a negative effect on the results when it reached 5, which is the common and default value for training word embedding models in many researches. Thus, this parameter must be carefully selected for the corpora. Probably the reason for the decrease in the accuracy of the results to the minimum count of 5 is due to the entry of rare words or typographical errors in the model training process, which leads to divergence in the generated models.

[14] Some selected models, which have been produced from hmBlogs can be accessed at http://fs.nlp.sbu.ac.ir/members/motahari/metr/papers/hmBlogs/models.

According to the above mentioned results and considerations and computational constraints for model training, the following hyperparameters were selected as the selected hyperparameters of the model training:

- Window Size = 5
- Using Skip-gram
- Dimension size = 400
- Min. Count = 100

It should also be mentioned that the number of epochs in all trained models was assumed to be 15. More detailed results for the trained Word2Vec models with the selected hyperparameters are given in the Table 7:

Table 7. Detailed accuracy percentage results of Word2Vec training for both hmBlogs and irBlogs with selected hyperparameters.

| Category | Number of Items | hmBlogs (Top-1) | irBlogs (Top-1) | hmBlogs (Top-5) | irBlogs (Top-5) | hmBlogs (Top-10) | irBlogs (Top-10) |
|---|---|---|---|---|---|---|---|
| Adjective-Adverb | 1332 | 13.964 | 10.21 | 31.081 | 25.375 | 40.541 | 32.432 |
| Antonym | 1260 | 22.619 | 20.079 | 51.19 | 47.063 | 58.254 | 53.333 |
| Country-Capital | 5402 | 26.194 | 16.216 | 45.428 | 32.469 | 52.388 | 39.559 |
| Province-Capital | 7832 | 30.209 | 24.119 | 46.246 | 41.254 | 50.958 | 46.221 |
| Comparative | 2520 | 26.667 | 21.429 | 63.413 | 48.413 | 71.825 | 52.778 |
| Currency | 1260 | 5.794 | 4.127 | 10.635 | 8.095 | 13.571 | 10.635 |
| Family Relationship | 342 | 50.585 | 44.737 | 71.93 | 65.497 | 82.164 | 73.099 |
| 1st Person | 1260 | 81.984 | 64.683 | 90.079 | 76.111 | 91.349 | 77.778 |
| Nationality | 1406 | 55.334 | 40.398 | 75.533 | 66.501 | 79.73 | 72.119 |
| Noun-Adverb | 1056 | 2.557 | 2.178 | 10.89 | 8.996 | 17.708 | 13.163 |
| Infinitive-Past | 1260 | 50.873 | 44.603 | 71.429 | 65.317 | 77.063 | 70.556 |
| Singular-Plural | 2550 | 37.804 | 35.216 | 73.059 | 70.431 | 82.824 | 81.333 |
| Infinitive-Present | 1260 | 35.159 | 18.095 | 66.667 | 48.016 | 77.857 | 60.079 |
| Superlative | 1260 | 21.27 | 16.111 | 48.73 | 43.175 | 58.73 | 50.952 |
| 3rd Person | 1332 | 69.144 | 58.033 | 84.459 | 77.928 | 89.64 | 81.682 |
| Total | 31332 | 32.698 | 25.434 | 53.514 | 45.503 | 60.047 | 51.57 |

As can be seen, hmBlogs has dominated irBlogs in all categories and all tests, and the divide is especially noticeable in many cases. Also some results of hmBlogs have very high scores, which is considered high accuracy for such experiments.

The selected hyperparameters were used in the next step to train models using the GloVe and FastText methods for the two competing corpora, hmBlogs and irBlogs. Also the Word2Vec models were created for the FaWiki and PRT corpora. Since the FaWiki corpus is a small corpus compared with the other corpora in this experiment, it was trained in two modes with Min. Count of 100 and 5. The results (only that of Top-5, due to display limitations) are shown in the Table 8 (the hmBlogs-W2V (Top-5) column in Table 8 is actually a repetition of the results from Table 7 and is mentioned again to help in the comparison):

Table 8. Detailed accuracy percentage results of models training on all corpora.

| Category | hmBlogs-W2V (Top-5) | hmBlogs-GloVe (Top-5) | irBlogs-GloVe (Top-5) | hmBlogs-FastText (Top-5) | irBlogs-FastText (Top-5) | PRT-W2V (Top-5) | FaWiki-W2V (Top-5) | FaWiki-W2V-Min. Count 5 (Top-5) |
|---|---|---|---|---|---|---|---|---|
| Adjective-Adverb | 31.081 | 28.604 | 14.94 | 31.607 | 31.306 | 18.093 | 12.688 | 9.084 |
| Antonym | 51.19 | 41.111 | 28.175 | 53.254 | 54.603 | 29.127 | 21.19 | 15.0 |
| Country-Capital | 45.428 | 37.338 | 16.327 | 48.501 | 38.06 | 50.833 | 56.498 | 56.646 |
| Province-Capital | 46.246 | 42.416 | 25.115 | 52.809 | 51.124 | 55.031 | 43.144 | 47.689 |
| Comparative | 63.413 | 38.333 | 27.143 | 74.206 | 65.952 | 56.984 | 17.46 | 20.873 |
| Currency | 10.635 | 11.825 | 7.063 | 10.159 | 8.81 | 12.857 | 9.127 | 7.302 |
| Family Relationship | 71.93 | 79.532 | 66.959 | 69.006 | 70.76 | 43.567 | 23.977 | 38.304 |
| 1st Person | 90.079 | 78.492 | 63.254 | 94.762 | 91.27 | 64.603 | 5.794 | 14.524 |
| Nationality | 75.533 | 69.63 | 44.523 | 77.454 | 75.605 | 49.004 | 78.378 | 75.178 |
| Noun-Adverb | 10.89 | 8.428 | 4.451 | 9.091 | 10.417 | 4.924 | 6.25 | 5.208 |
| Infinitive-Past | 71.429 | 69.206 | 55.714 | 71.746 | 72.063 | 48.016 | 34.048 | 49.921 |
| Singular-Plural | 73.059 | 61.294 | 48.353 | 70.51 | 71.765 | 35.961 | 42.824 | 44.824 |
| Infinitive-Present | 66.667 | 50.0 | 28.571 | 66.111 | 47.54 | 49.286 | 4.841 | 9.365 |
| Superlative | 48.73 | 35.873 | 28.333 | 56.667 | 58.651 | 38.73 | 28.175 | 28.81 |
| 3rd Person | 84.459 | 74.7 | 63.438 | 89.94 | 87.237 | 72.898 | 49.324 | 70.195 |
| Total | 53.514 | 45.302 | 29.912 | 57.156 | 53.431 | 46.496 | 36.19 | 39.375 |

The results of the Table 8 show lower uniformity than that of the Table 7 and indicate some unknown behavior of the models and corpora. The following conclusions can be made:

- In terms of overall performance, hmBlogs still performs better than all other corpora.
- GloVe performed significantly lower than Word2Vec.
- The GloVe model of irBlogs had the worst results compared with hmBlogs (over 15% distance in accuracy score). IrBlogs showed most vulnerability when it was used by GloVe.
- Using FastText enabled irBlogs to surpass hmBlogs in 5 categories by a thin margin for the first time. The use of Word2Vec and GloVe enabled hmBlogs to dominate in all the test cases, and the use of FastText helped irBlogs narrow its distance with hmBlogs.
- Despite being significantly larger than hmBlogs and irBlogs in terms of word count, PRT failed to reach hmBlogs and performed slightly better than irBlogs (see irBlogs functionality based on Word2Vec in the Table 7). In fact, with Word2Vec, PRT performed almost similar to irBlogs.
- While the Persian wiki corpus (FaWiki) is used in many researches for word embedding training, it yielded much worse results than the other corpora used in this experiment. Although, it did perform exceptionally well for some categories such as the Nationality or Country-Capital category than the other corpora. This suggests that while this corpus may be far from the common language of Persian-speakers, it contains specific, statistical or specialized data that can make it applicable in such fields.
- Decreasing the Min. Count from 100 to 5 did not really affect the quality of the FaWiki models. Interestingly, the highest score of the Nationality category was obtained from the FaWiki Word2Vec model, which is based on the Min. Count 100 (relative to Min. Count 5). And this again indicates that different hyperparameters may be needed to meet the specific application, and no general rule can be made in this regard.

**4-3-2- Experiment on FATS**

Another dataset that is used for evaluation is FATS. Due to the size of the dataset and the limitations on the hardware computational power, the FaWiki, which had the weakest scores in the previous evaluation, was removed from the comparison list, and the evaluation was made for only three corpora: hmBlogs, irBlogs and PRT and was based on the Word2Vec models, which are trained by selected hyperparameters. Table 9 illustrates a comparison of the models with all three datasets created in 4-2-2. Due to display limitations, the table only shows a summary of the results; a more detailed report of the outputs is shown in the Table 10.

Table 9. The summary of accuracy percentage of the FATS experiment.

| FATS Dataset | hmBlogs (Top-1) | irBlogs (Top-1) | PRT (Top-1) | hmBlogs (Top-5) | irBlogs (Top-5) | PRT (Top-5) | hmBlogs (Top-10) | irBlogs (Top-10) | PRT (Top-10) |
|---|---|---|---|---|---|---|---|---|---|
| FATS-100 | 1.654 | 1.349 | 0.880 | 4.915 | 3.898 | 2.497 | 7.322 | 5.606 | 3.579 |
| FATS-250 | 1.663 | 1.348 | 0.945 | 4.936 | 3.839 | 2.594 | 7.190 | 5.472 | 3.759 |
| FATS-500 | 1.908 | 1.656 | 1.238 | 5.178 | 4.250 | 2.969 | 7.416 | 5.964 | 4.132 |

Full details of the experiment for all categories of FATS-500 and Top-5 are shown in the Table 10:

Table 10. Details of the accuracy percentage of the FATS-500 experiment.

| Category | Number of Items | hmBlogs (Top-1) | irBlogs (Top-1) | PRT (Top-1) | hmBlogs (Top-5) | irBlogs (Top-5) | PRT (Top-5) | hmBlogs (Top-10) | irBlogs (Top-10) | PRT (Top-10) |
|---|---|---|---|---|---|---|---|---|---|---|
| Agent | 72 | 8.333 | 5.556 | 2.778 | 16.667 | 12.5 | 5.556 | 22.222 | 19.444 | 5.556 |
| Antonym | 249500 | 5.82 | 4.371 | 4.77 | 12.438 | 8.885 | 7.725 | 15.891 | 11.174 | 9.32 |
| Attribute | 1980 | 0.051 | 0.303 | 0.101 | 0.404 | 1.465 | 0.303 | 1.162 | 2.626 | 0.556 |
| Cause | 56 | 1.786 | 0.0 | 0.0 | 5.357 | 1.786 | 1.786 | 5.357 | 1.786 | 1.786 |
| Domain | 249500 | 0.264 | 0.238 | 0.146 | 1.505 | 1.049 | 0.556 | 2.62 | 1.793 | 0.871 |
| Has-Potential defining feature | 6 | 0.0 | 0.0 | 0.0 | 0.0 | 0.0 | 0.0 | 0.0 | 0.0 | 0.0 |
| Has-Salient defining feature | 870 | 0.0 | 0.115 | 0.115 | 2.874 | 0.92 | 0.69 | 5.402 | 2.184 | 1.494 |
| Has-Unit | 42 | 0.0 | 0.0 | 0.0 | 0.0 | 0.0 | 0.0 | 0.0 | 0.0 | 0.0 |
| Hypernym | 249500 | 0.539 | 0.463 | 0.286 | 1.908 | 1.433 | 1.127 | 3.141 | 2.141 | 1.715 |
| Hyponym | 249500 | 0.483 | 0.283 | 0.202 | 2.164 | 1.6 | 1.032 | 3.702 | 2.961 | 1.808 |
| Instance hypernym | 249500 | 2.23 | 2.782 | 0.552 | 5.518 | 6.576 | 2.369 | 7.755 | 8.874 | 3.707 |
| Instance hyponym | 1560 | 0.513 | 0.449 | 0.192 | 4.038 | 2.949 | 0.897 | 6.987 | 6.795 | 1.41 |
| Instrument | 56 | 0.0 | 0.0 | 0.0 | 0.0 | 0.0 | 0.0 | 0.0 | 0.0 | 0.0 |
| Is-Agent-of | 342 | 2.047 | 3.216 | 0.877 | 9.942 | 8.48 | 4.386 | 14.912 | 10.526 | 5.848 |
| Is-Attribute-of | 98910 | 0.845 | 0.646 | 0.23 | 3.682 | 2.722 | 1.224 | 5.432 | 4.093 | 2.276 |
| Is-Caused-by | 42 | 11.905 | 14.286 | 4.762 | 26.19 | 21.429 | 11.905 | 35.714 | 30.952 | 23.81 |
| Is-Domain-of | 1056 | 0.0 | 0.095 | 0.0 | 0.284 | 0.473 | 0.379 | 1.136 | 0.852 | 0.568 |
| Is-Instrument-of | 930 | 2.688 | 0.753 | 0.538 | 9.14 | 5.699 | 3.118 | 13.333 | 10.538 | 5.269 |
| Is-Location-of | 2 | 50.0 | 0.0 | 0.0 | 50.0 | 50.0 | 0.0 | 50.0 | 50.0 | 50.0 |
| Is-Patient-of | 1722 | 2.149 | 1.336 | 0.987 | 5.865 | 4.646 | 5.749 | 9.117 | 6.794 | 8.653 |
| Location | 90 | 0.0 | 0.0 | 0.0 | 1.111 | 1.111 | 0.0 | 2.222 | 1.111 | 0.0 |
| Member holonym | 23256 | 2.391 | 1.84 | 0.83 | 7.856 | 6.308 | 3.212 | 11.601 | 9.064 | 4.911 |
| Member meronym | 19182 | 0.563 | 0.511 | 0.162 | 3.045 | 2.299 | 1.001 | 5.453 | 3.952 | 1.971 |
| Part holonym | 249500 | 3.322 | 3.228 | 2.387 | 9.094 | 8.166 | 5.927 | 12.703 | 11.051 | 8.469 |
| Part meronym | 249500 | 1.299 | 0.891 | 0.928 | 4.153 | 2.975 | 2.853 | 6.576 | 4.811 | 3.814 |
| Patient | 1722 | 0.058 | 0.174 | 0.058 | 0.813 | 1.103 | 0.174 | 1.974 | 1.974 | 0.174 |
| Portion holonym | 182 | 0.549 | 0.0 | 0.0 | 1.648 | 1.648 | 0.0 | 4.945 | 3.846 | 0.0 |
| Portion meronym | 42 | 0.0 | 0.0 | 0.0 | 0.0 | 0.0 | 0.0 | 0.0 | 0.0 | 0.0 |
| Potential defining feature | 20880 | 2.112 | 1.82 | 1.571 | 6.705 | 4.248 | 3.487 | 9.219 | 5.551 | 4.397 |
| Salient defining feature | 552 | 0.725 | 0.725 | 0.181 | 3.08 | 2.536 | 0.725 | 8.152 | 3.986 | 2.355 |
| Substance holonym | 4970 | 1.69 | 0.523 | 1.529 | 5.714 | 2.093 | 4.286 | 8.954 | 3.441 | 5.453 |
| Substance meronym | 32220 | 1.251 | 0.59 | 0.636 | 4.544 | 2.291 | 2.976 | 7.194 | 3.582 | 4.69 |
| Unit | 72 | 0.0 | 0.0 | 0.0 | 0.0 | 0.0 | 0.0 | 0.0 | 1.389 | 1.389 |
| Total | 1957314 | 1.908 | 1.656 | 1.238 | 5.178 | 4.25 | 2.969 | 7.416 | 5.964 | 4.132 |

As evident in this experiment, hmBlogs has also won over its competitors. However, the accuracy percentages are significantly lower than those observed in the previous dataset (see 4-3-1). This finding is noteworthy from various aspects:

- The semantic categories in FarsNet are not balanced in regard to the number of relations. For example, the total number of relations in FarsNet is less than 10 for several categories, such as 'Agent', 'Cause', 'Unit', and 'Portion meronym'. This issue can disrupt uniformity and cause an imbalance in dataset production. As can be seen some categories have less than 100 members while some others have more than 240,000 members.
- Since analogy questions are generated based on the participation of each word of a FarsNet sense in a relation, polysemous words can disrupt the overall results. In fact in the case of polysemous words, models may tend only to the dominant (i.e. frequent) sense of the words, because models are not trained to consider context. So the use of rare and infrequent senses of a word in a relationship may naturally lead to erroneous results.
- Since only a small number of relations are randomly selected due to computational constraints, it is quite possible that the results will be improved if more relations are used. As shown in the Table 9, increasing the number of relationships (in FATS-500 versus FATS-250 and FATS-100) gradually and slightly improves the accuracy of the results.
- And while word embedding models are slightly more complex and better understand certain relations, such as the 'Is-Caused-By', they do not provide tangible results for some other relations (such as 'Unit'). In fact, it can be said that some semantic relationships are indistinguishable from word embedding models, at least by the current training methods.

- Because the accuracy percentage is obtained from the aggregation of the results of all the questions, very low results from certain categories of the experiment strongly affects the overall results and reduces the overall accuracy percentage. This is especially true about the categories with the highest number of questions.

**4-3-3- Experiment on Semantic Relatedness**

One of the methods to evaluate word embedding models is to use words semantic relatedness. As mentioned in 4-2-3, SemEval2017-task2 datasets are used for this purpose. Table 11 illustrates the correlation between the values in the dataset and similarity outputs provided by the models of this study:

Table 11. Results on Pearson correlation coefficient calculation between SemEval2017-task2 and models of the experiment.

| Pearson correlation coefficient on SemEval2017 | hmBlogs-W2V | irBlogs-W2V | hmBlogs-GloVe | irBlogs-GloVe | hmBlogs-FastText | irBlogs -FastText | PRT-W2V | FaWiki-W2V | FaWiki-W2V-Min. Count 5 |
|---|---|---|---|---|---|---|---|---|---|
| Value | 0.571 | 0.459 | 0.540 | 0.480 | 0.679 | 0.638 | 0.549 | 0.385 | 0.456 |

As observed in the table, hmBlogs continues to remain at the top. Furthermore, the Word2Vec model of PRT was more successful than the irBlogs model, in comparison with that of previous experiments (see 4-3-1 and 4-3-2). Better performance of the PRT over irBlogs could be due to the dataset or the test itself, which again demonstrates the importance of datasets, parameters, training methods, evaluation methods and corpus types in results. Nevertheless, hmBlogs maintained its domination throughout all the different circumstances.

**5- Conclusion**

This paper aimed to introduce a large, open and general corpus for the Persian language, as a low resource language. The introduced corpus is, by far, the largest of all the few independent and publicly available Persian corpora (with the exception of PRT, which is not an independent corpus but a collection of several corpora). The hmBlogs corpus is a rather large corpus even when compared to the English corpora (for example see (*English Corpora*, n.d.)). HmBlogs has had the latest style of Persian writing in recent years and has preserved texts from the last 15 years in the Persian blog space.

HmBlogs showed overall better performance than other corpora such as irBlogs, PRT and FaWiki in word embedding models. A new analogy dataset, (FATS), was also presented as a side product of this research along with the obtained models. Also all factors including the corpora, evaluation data, evaluation methods, model construction methods and hyperparameters were found to have notable effects on the evaluation output.

One of the challenges is keeping such corpora up to date. To solve this problem, it may be necessary to design processes that can keep corpora with the latest changes and adding new content. Designing such a process and its related systems could be a topic for future studies. The hmBlogs can also be considered for further development with new resources, such as crawling and adding post comments, in the future. Another possibility could be combining the corpus with other sources, such as news sources, the Persian wiki, and social media data.

Tokenization poses another challenge to the Persian language because many words in Persian are multi-token. Even in some cases, particularly in some Persian verbs, tokens of a word are not consecutive and many words can fall between the gaps. Analyzing this phenomenon was beyond the scope of this research and should be addressed independently, but further research in this area can help to increase the quality of language models produced from corpora.

A further task can also be the labeling of the present corpus. Topic modeling, official/informal/colloquial language labeling, tagging the polarity of posts, and annotating highly similar or nearly duplicated contents are all topics for further studies.